\PassOptionsToPackage{unicode}{hyperref}
\PassOptionsToPackage{hyphens}{url}
\documentclass[
  english,
  ]{article}
\usepackage{amsmath,amssymb}
\usepackage{lmodern}
\usepackage{ifxetex,ifluatex}
\ifnum 0\ifxetex 1\fi\ifluatex 1\fi=0 
  \usepackage[T1]{fontenc}
  \usepackage[utf8]{inputenc}
  \usepackage{textcomp} 
\else 
  \usepackage{unicode-math}
  \defaultfontfeatures{Scale=MatchLowercase}
  \defaultfontfeatures[\rmfamily]{Ligatures=TeX,Scale=1}
\fi
\IfFileExists{upquote.sty}{\usepackage{upquote}}{}
\IfFileExists{microtype.sty}{
  \usepackage[]{microtype}
  \UseMicrotypeSet[protrusion]{basicmath} 
}{}
\makeatletter
\@ifundefined{KOMAClassName}{
  \IfFileExists{parskip.sty}{%
    \usepackage{parskip}
  }{
    \setlength{\parindent}{0pt}
    \setlength{\parskip}{6pt plus 2pt minus 1pt}}
}{
  \KOMAoptions{parskip=half}}
\makeatother
\usepackage{xcolor}
\IfFileExists{xurl.sty}{\usepackage{xurl}}{} 
\IfFileExists{bookmark.sty}{\usepackage{bookmark}}{\usepackage{hyperref}}
\hypersetup{
  pdftitle={DHSEGATs: Distance and Hop-wise Structures Encoding Enhanced Graph Attention Networks},
  pdflang={en},
  pdfkeywords={Graph Attention Networks, Graph Structure
Information, Label Propagation},
  hidelinks,
  pdfcreator={LaTeX via pandoc}}
\usepackage[a4paper,bindingoffset=0mm,inner=30mm,outer=30mm,top=30mm,bottom=30mm]{geometry}
\usepackage{longtable,booktabs,array}
\usepackage{calc} 
\usepackage{etoolbox}
\makeatletter
\patchcmd\longtable{\par}{\if@noskipsec\mbox{}\fi\par}{}{}
\makeatother
\IfFileExists{footnotehyper.sty}{\usepackage{footnotehyper}}{\usepackage{footnote}}
\makesavenoteenv{longtable}
\usepackage{graphicx}
\makeatletter
\def\maxwidth{\ifdim\Gin@nat@width>\linewidth\linewidth\else\Gin@nat@width\fi}
\def\maxheight{\ifdim\Gin@nat@height>\textheight\textheight\else\Gin@nat@height\fi}
\makeatother
\setkeys{Gin}{width=\maxwidth,height=\maxheight,keepaspectratio}
\makeatletter
\def\fps@figure{htbp}
\makeatother
\providecommand{\tightlist}{%
  \setlength{\itemsep}{0pt}\setlength{\parskip}{0pt}}
\ifxetex
  \usepackage{polyglossia}
  \setmainlanguage[]{english}
\else
  \usepackage[main=english]{babel}

\def\languageshorthands#1{}
\fi
\ifluatex
  \usepackage{selnolig}  
\fi
\newlength{\cslhangindent}
\newlength{\csllabelwidth}
\newenvironment{CSLReferences}[2] 
 {
  \setlength{\parindent}{0pt}
  \ifodd #1 \everypar{\setlength{\hangindent}{\cslhangindent}}\ignorespaces\fi
  \ifnum #2 > 0
  \setlength{\parskip}{#2\baselineskip}
  \fi
 }%
 {}
\usepackage{calc}

\newcommand{\CSLLeftMargin}[1]{\parbox[t]{\csllabelwidth}{#1}}
\newcommand{\CSLRightInline}[1]{\parbox[t]{\linewidth - \csllabelwidth}{#1}\break}

\title{DHSEGATs: Distance and Hop-wise Structures Encoding Enhanced
Graph Attention Networks}

\author{
            Zhiguo Huang \\
        Sci-Tech Academy of ZheJiang University;Research Center of
Hundsun LTD. \\
        Hangzhou, China, 310030 \\
        hzg0601@163.com \\
         \and
            Xiaowei Chen \\
        School of Finance, NanKai University \\
        Tianjin, China, 300381 \\
        chenx@nankai.edu.cn \\
         \and
            Bojuan Wang \\
        School of Finance, NanKai University \\
        Tianjin, China, 300381 \\
        jphsb@163.com \\
        }

\date{}

\begin{document}
\maketitle
\begin{abstract}
Numerous works have proven that existing neighbor-averaging Graph Neural
Networks cannot efficiently catch structure features, and many works
show that injecting structure, distance, position or spatial features
can significantly improve performance of GNNs, however, injecting
overall structure and distance into GNNs is an intuitive but remaining
untouched idea. In this work, we shed light on the direction. We first
extracting hop-wise structure information and compute distance
distributional information, gathering with node's intrinsic features,
embedding them into same vector space and then adding them up. The
derived embedding vectors are then fed into GATs(like GAT, AGDN) and
then Correct and Smooth, experiments show that the DHSEGATs achieve
competitive result. The code is available at
https://github.com/hzg0601/DHSEGATs.
\end{abstract}

\hypertarget{introduction}{%
\subsection{1. Introduction}\label{introduction}}

Many works have proven that existing neighbor-averaging Graph Neural
Networks cannot efficiently catch structure information, such GNNs
cannot even catch degree features in some cases. The reason is
intuitive: as the neighbor-averaging GNNs can only combine neighbor's
feature vectors for every node, if the neighbor's feature vectors
contains no structure information, the hop-wise neighbor-averaging GNNs
can only catch degree information at
best(\protect\hyperlink{ref-li2018deeper}{{[}1{]}};\protect\hyperlink{ref-oono2019graph}{{[}2{]}};\protect\hyperlink{ref-dehmamy2019understanding}{{[}3{]}}).
So, as an intuitive idea, injecting structure information into feature
vectors may improve the performance of GNNs.

Numerous works have shown that injecting structure, distance, position
or spatial information can significantly improve performance of
neighbor-averaging
GNNs(\protect\hyperlink{ref-monti2018motifnet}{{[}4{]}};\protect\hyperlink{ref-bouritsas2020improving}{{[}5{]}};\protect\hyperlink{ref-you2021identityaware}{{[}6{]}};\protect\hyperlink{ref-ying2021transformers}{{[}7{]}};\protect\hyperlink{ref-you2019positionaware}{{[}8{]}};\protect\hyperlink{ref-li2020distance}{{[}9{]}};\protect\hyperlink{ref-alsentzer2020subgraph}{{[}10{]}}).
However, existing works have their problems. Some of them has very high
computation complexity which can not apply to large-scale
graph(MotifNet\protect\hyperlink{ref-monti2018motifnet}{{[}4{]}}). Some
of them simply concatenate structure information with intrinsic feature
vector (ID-GNN\protect\hyperlink{ref-you2021identityaware}{{[}6{]}};
P-GNN\protect\hyperlink{ref-you2019positionaware}{{[}8{]}};
DE-GNN\protect\hyperlink{ref-li2020distance}{{[}9{]}}), which may
confuse the signals of different feature. For example, in ogbn-arxiv
dataset, the intrinsic feature is semantic embedding of headline or
abstract, which provides total different signal with structure
information. Some of them are graph-level-task oriented and only deal
with small
graph(Graphormer\protect\hyperlink{ref-ying2021transformers}{{[}7{]}} ;
SubGNN\protect\hyperlink{ref-alsentzer2020subgraph}{{[}10{]}}).
Moreover, existing work considerate only simple structure information
such as degree and circles, remaining injecting high-level structure
information untouched. More importantly, hop-wise neighbor-averaging
GNNs' sensitive field is restricted into only one hop and beyond to
catch multi-hop structure information. To inject hop-wise structure
information also remain untouched.

In this work, we shed light on how efficiently injecting high-level
multi-hop structure and distance information into GATs. By investigating
computation complexity of popular node-level and graph-level structural
information, we propose a \(O(|N|)\) computation complexity combination
of structural information indicators. For every node, we first extract k
hop-wise ego-nets to compute node-level and graph-level structure
information indicators. At the same time, a distance sequence of k-hop
ego-net is extracted to compute distributional features. The distance,
hop-wise structure information and node's intrinsic features are encoded
into the same vector space, just like transformer's initial embeddings.
Derived feature vectors are fed into GATs to get initial predictions.
Those initial predictions are then fed into Correct and Smooth to inject
global label information. Experiments show that the scheme can
significantly improve the performance of GATs.

our contribution are concluded as follow:

\begin{enumerate}
\def\labelenumi{\arabic{enumi}.}
\tightlist
\item
  By investigating popular structure indicators computation complexity,
  we propose a structure indicator combination with \(O(|N|)\)
  computation complexity, and propose a distance distributional
  information extracting scheme that does not corrupt graph's orginal
  structure.
\item
  we propose a hop-wise high-level distance and structure information
  injecting scheme which fits universal neighor averaing GNN models.
\item
  we conduct extensive experiments to demonstrate the effectiveness of
  injecting distance and hop-wise structure information into GATs.
\end{enumerate}

\hypertarget{related-works}{%
\subsection{2. Related Works}\label{related-works}}

\hypertarget{structure-and-distance-information-in-gnns}{%
\subsubsection{2.1 Structure and Distance Information in
GNNs}\label{structure-and-distance-information-in-gnns}}

\begin{enumerate}
\def\labelenumi{\arabic{enumi}.}
\tightlist
\item
  \textbf{Structure Information}
\end{enumerate}

ID-GNN\protect\hyperlink{ref-you2021identityaware}{{[}6{]}} argues that
without external features, GNNs cannot outperform 1-WL test, however,
injecting number of circles makes GNNs more powerful. In-degree and
out-degree embedding is an integral part of Graphormer
(\protect\hyperlink{ref-ying2021transformers}{{[}7{]}}) for winning OGB
Large-Scale Challenge 2021. By injecting number of motifs in graph,
MotifNet(\protect\hyperlink{ref-monti2018motifnet}{{[}4{]}})makes
significant improvement on enlarging GNNs expressive power. By subgraph
isomorphism counting ,
GSN(\protect\hyperlink{ref-bouritsas2020improving}{{[}5{]}}) enlarges
the expressive power of
GNNs;\protect\hyperlink{ref-dasoulas2019coloring}{{[}11{]}} proposes
Colored Local Iterative Procedure(CLIP) to inject structure information
into GNNs, which also improve GNNs performance.

\begin{enumerate}
\def\labelenumi{\arabic{enumi}.}
\setcounter{enumi}{1}
\tightlist
\item
  \textbf{Position, Distance and Spatial Information}
\end{enumerate}

By injecting spatial information into transformer,
Graphormer(\protect\hyperlink{ref-ying2021transformers}{{[}7{]}})wins
OGB Large-Scale Challenge 2021 and make transformer be the best model in
graph classification. By injecting position information,
P-GNN(\protect\hyperlink{ref-you2019positionaware}{{[}8{]}}) achieve
significant improvement comparing with baseline GNNs. By injecting
distance encoding information,
DE-GNN(\protect\hyperlink{ref-li2020distance}{{[}9{]}})outperm baseline
GNNs . By injecting neighborhood, structure and position information,
SubGNN(\protect\hyperlink{ref-alsentzer2020subgraph}{{[}10{]}})achieve
SOTA performance on subgraph classification.

\hypertarget{the-disadvantage-of-neighbor-averaging-gnns}{%
\subsubsection{2.2 The Disadvantage of Neighbor Averaging
GNNs}\label{the-disadvantage-of-neighbor-averaging-gnns}}

Early works that analyze disadvantage of GNNs comes from the aspect of
oversmoothing problem.
\protect\hyperlink{ref-li2018deeper}{{[}1{]}}analyzes the asymptotic
behavior of oversmoothing and finds that when without nonlinear
activation function, GNNs can only catch degree information.
\protect\hyperlink{ref-oono2019graph}{{[}2{]}}makes further efforts and
demonstrates that even with nonlinear activation function, GNNs can also
only catch degree
information.\protect\hyperlink{ref-xu2018how}{{[}12{]}} demonstrates
that message passing GNNs cannot outperform 1-WL in aspect of expressive
power. \protect\hyperlink{ref-dehmamy2019understanding}{{[}3{]}}shows
that without proper convolution support, message passing GNN cannot even
catch degree information.

Another idea to analyze disadvantage of GNNs is to inspect the ability
of GNNs to count simple local structure indicators such as circles and
triangles. Their finds are consistent with
GIN\protect\hyperlink{ref-xu2018how}{{[}12{]}}, which can basically be
concluded as: message passing GNNs can not count local structure
features that 1-WL test can not
count(\protect\hyperlink{ref-bouritsas2020improving}{{[}5{]}},
\protect\hyperlink{ref-arvind2020weisfeilerleman}{{[}13{]}}--\protect\hyperlink{ref-vignac2020building}{{[}15{]}}).
Some works inspect the GNNs' spectral ability and find that GNNs are
nothing but low-pass
filters\protect\hyperlink{ref-nt2019revisiting}{{[}16{]}}.
SIGN(\protect\hyperlink{ref-frasca2020sign}{{[}17{]}})and
SAGN(\protect\hyperlink{ref-sun2021scalable}{{[}18{]}}) show that by
concatenating graph diffusion processed features , MLP outperforms most
of complicate GNNs, demonstrating the disadvantage of GNNs in another
way.

\hypertarget{graph-attention-mechanism}{%
\subsubsection{2.3 Graph Attention
Mechanism}\label{graph-attention-mechanism}}

The reason why we prefer GATs among large amounts of GNN models is that
is has two advantages:

1, According to GIN(\protect\hyperlink{ref-xu2018how}{{[}12{]}}), the
key of expressive power of a message passing GNN is that its aggregating
function is multi-set injective or not, while the attention mechanism of
GATs intrinsically provides multi-set injective aggregating function.

2, GNNs are troubled by the problem of over-smoothing, however, because
of residual connection and attention mechanism, GATs can theoretically
get rid of over-smoothing and make number of layers go deeper.

Because of the two advantages, attention mechanism on graph arouses
enormous interest. Representative works including:
GAT(\protect\hyperlink{ref-velickovic2017graph}{{[}19{]}}),
GTN(\protect\hyperlink{ref-wu2020graph}{{[}20{]}}),
GPT-GNN(\protect\hyperlink{ref-hu2020gptgnn}{{[}21{]}}),
HGT(\protect\hyperlink{ref-hu2020heterogeneous}{{[}22{]}}),
Graphormer(\protect\hyperlink{ref-ying2021transformers}{{[}7{]}}),
Graphormer(\protect\hyperlink{ref-ying2021transformers}{{[}7{]}}),
AGDN(\protect\hyperlink{ref-sun2020adaptive}{{[}23{]}}).
GAT(\protect\hyperlink{ref-velickovic2017graph}{{[}19{]}}) is the first
work that combines message passing mechanism with attention mechanism.
After that, GTN(\protect\hyperlink{ref-wu2020graph}{{[}20{]}}) further
combines GNN with transformer,
GPT-GNN(\protect\hyperlink{ref-hu2020gptgnn}{{[}21{]}}) combines GNN
with GPT. HGT(\protect\hyperlink{ref-hu2020heterogeneous}{{[}22{]}})
adapt Graph Transformer for heterogeneous graph. Comparing with former
graph attention networks,
Graphormer(\protect\hyperlink{ref-ying2021transformers}{{[}7{]}})adopts
full transformer and get rids of message passing mechanism.
AGDN(\protect\hyperlink{ref-sun2020adaptive}{{[}23{]}}) combines graph
attention mechanism and graph diffusion process which outperforms most
of existing single GNN model on ogbn-arxiv dataset.

\hypertarget{label-propagation-in-gnns}{%
\subsubsection{2.4 Label Propagation in
GNNs}\label{label-propagation-in-gnns}}

Label propagation in graph has been proven a effective way to inject
global label information in the context of semi-supervised learning. Its
basic assumption is that labels of adjacent nodes has higher probability
to be same than nodes are not adjacent. Recently, large amounts of work
concentrate on combining GNN with label propagation. Some works combine
label propagation with GNN directly such as taking label propagation as
extral loss(\protect\hyperlink{ref-wang2020unifying}{{[}24{]}}). Some
works inject label information by Markov Random
Field(\protect\hyperlink{ref-qu2019gmnn}{{[}25{]}},
\protect\hyperlink{ref-gao2019conditional}{{[}26{]}}). Some works take
label information as extra
feature(\protect\hyperlink{ref-zheng2021gipa}{{[}27{]}}). And some
others first do error correction of label prediction of base model and
then use the error correction to smooth prediction like regression and
get surprisely good performance even combining with
MLP(\protect\hyperlink{ref-klicpera2018predict}{{[}28{]}},
\protect\hyperlink{ref-huang2020combining}{{[}29{]}}).

\hypertarget{methodology}{%
\subsection{3. Methodology}\label{methodology}}

\hypertarget{overall-architecture}{%
\subsubsection{3.1 Overall Architecture}\label{overall-architecture}}

The overall architecture of our model is depicted as follow: First,
extracting \(k\) hop-wise ego-nets of target node then computing
node-level and graph-level structure information indicators. Besides,
in-degree and out-degree also are extracted as a kind of structure
information. At the same time, computing distance of every node of
\(k-hop\) ego-net to target node as a distance sequence. Then computing
distributional indicators of the distance sequence as distance encoding
vectors. The distance encoding vectors, hop-wise structure information
and node's intrinsic features are encoded into the same vector space
then adding up together, just like transformer's initial embeddings.
Derived feature vectors are fed into GATs to get soft predictions. Those
soft predictions are then fed into Correct and Smooth to inject global
label information to get final predictions.

\begin{figure}
\centering
\includegraphics{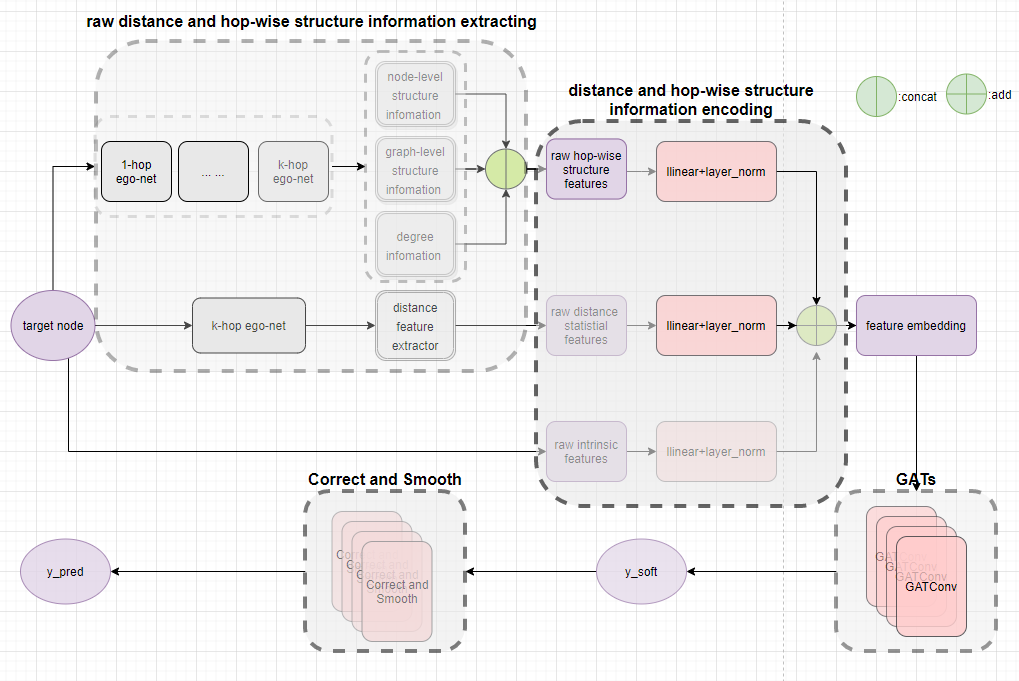}
\caption{architecture}
\end{figure}

Figure 1 overall architecture of DHSEGATs

\hypertarget{preprocessing}{%
\subsubsection{3.2 Preprocessing}\label{preprocessing}}

\hypertarget{degree-features}{%
\paragraph{3.2.1 Degree Features}\label{degree-features}}

Drawing lessons from
Graphormer(\protect\hyperlink{ref-ying2021transformers}{{[}7{]}})
embeding in-degree and out-degree as centrality feature, we also use
in-degree and out-degree but as a part of structure information.

\hypertarget{hop-wise-ego-nets-node-level-and-graph-level-features}{%
\paragraph{3.2.2 Hop-wise Ego-net's Node-level and Graph-level
Features}\label{hop-wise-ego-nets-node-level-and-graph-level-features}}

As mentioned before, neighbor-averaging GNNs can only catch degree
information at best, and degree information is merely one-hop node-level
structure information. To catch multi-hop structure information, we
first extract \(k\) \(r\)-hop ego-net(\(r=1,...,k\)), then compute their
node-level and graph-level structure information. However, many popular
structure information has very high computation complexity, for example,
the computation complexity of eigenvector centrality is \(O(|N|^3)\),
and betweenness centrality's is \(O(|N|*|E|)\)(where \(|E|\) denotes
number of edges). It's unrealistic to apply to large-scale graph if the
preprocessing has very high computation complexity. Therefore, we first
investigate computation complexity of popular structure indicators, from
which we choose following indicators with \(O(|N|)\) computation
complexity:

\begin{enumerate}
\def\labelenumi{\arabic{enumi}.}
\tightlist
\item
  \textbf{Node-level Structure Information: Triangles, Clustering,
  Square Clustering}
\end{enumerate}

Triangles: the number of triangles contains target node in ego-net.

Clustering: \(c_u = \frac{2 T(u)}{deg(u)(deg(u)-1)}\), where \(T(u)\) is
the number of triangles through node \(u\) and \(deg(u)\) is the degree
of \(u\).

Square-
clustering:\(C_4(v) = \frac{ \sum_{u=1}^{k_v} \sum_{w=u+1}^{k_v} q_v(u,w) }{ \sum_{u=1}^{k_v} \sum_{w=u+1}^{k_v} [a_v(u,w) + q_v(u,w)]}\),
where \$q\_v(u,w) \$are the number of common neighbors of \(u\) and
\(w\) other than \(v\) (ie squares), and
\(a_v(u,w) = (k_u - (1+q_v(u,w)+\theta_{uv})) + (k_w - (1+q_v(u,w)+\theta_{uw}))\),
where \(\theta_{uw} = 1\) if \(u\) and \(w\) are connected and \(0\)
otherwise.

\begin{enumerate}
\def\labelenumi{\arabic{enumi}.}
\setcounter{enumi}{1}
\tightlist
\item
  \textbf{Graph-level Structure Information: Density, Number of
  Self-loops, Transitivity}
\end{enumerate}

Density: \(d = \frac{2m}{n(n-1)},\) where \(n\) is the number of nodes
and \(m\) is the number of edges in graph.

Number of self-loops: number of self-loop edges, where a self-loop edge
has the same node at both ends.

Transitivity: \(T = 3\frac{\#triangles}{\#triads}\), where a triad is
graph contains of two edges with a shared vertex.

\hypertarget{distance-distributional-features}{%
\paragraph{3.2.3 Distance Distributional
Features}\label{distance-distributional-features}}

Distance information is a very important information for attention
mechanism. In order to get larger sensitive field of distance
information, we first extract \(k\)-hop ego-net arounds target node, and
compute every neighbor node's unweighted distance in the ego-net to
target node to get a distance sequence. However, we cannot directly
apply the sequence to GNNs because the sequence is variant and may be
very long. As a result, we should encoding it to a fixed length vector
beforehand. Technicallly, we can adopt neighbor sampling and padding
technique, but this way may corrupt graph's structure and introduce a
virtual node, which may result disastrous consequence in encoding. Hence
we adopt an encoding method based on distributional indicators, that is
to compute statistics of the sequence. We chose seven statistics,
including: maximum, minimum, median, mean, standard deviation, kurtosis,
skewness.

\hypertarget{encoding-distance-and-hop-wise-structure-features}{%
\subsubsection{3.3 Encoding Distance and Hop-wise Structure
Features}\label{encoding-distance-and-hop-wise-structure-features}}

In order to encode distance and hop-wise structure information, we first
encode distance distributional information ,hop-wise structure feature
,intrinsic feature to same vector space by three linear layers:

\begin{equation}
hd = dist\_encoder(raw \ distance \ distributional \ feature)
\end{equation} \begin{equation}
hs = struc\_encoder(raw \ hop-wise \ structure \ feature)
\end{equation} \begin{equation}
hi = intri\_encoder(raw \ intrinsic \ feature)
\end{equation}

In order to avoid one of the there feature dominate the initial encoding
vector, layer norm modules are followed: \begin{equation}
hi = layer\_norm(hi)
\end{equation} \begin{equation}
hd = layer\_norm(hd)
\end{equation} \begin{equation}
hs = layer\_norm(hs)
\end{equation}

In order to reduce the spatial complexity, the three encoded vectors are
added up and get initial encoding vector as follow: \begin{equation}
h = hi+hd+hs
\end{equation}

\hypertarget{gats}{%
\subsubsection{3.4 GATs}\label{gats}}

In this work, we investigate two GAT models: GAT and AGDN. The former is
most classical attention model on graph and a generally used baseline
model. The latter achieve best performance on ognb-arxiv dataset among
single models by combining attention mechanims and graph diffusion
process.

\hypertarget{gat}{%
\paragraph{3.4.1 GAT}\label{gat}}

A general form of neighbor-averaging GNN could be written as:
\begin{equation}
h_i^{\prime} = UPD_{W} \left( h_i,AGG_{j \in N(i)} \, MES_{W}
\left(h_i, h_j,e_{j,i}\right) \right)
\end{equation}

where \(h\), \(AGG\), \(MES\) , \(UPD\) denote nodes' representation,
aggregating function, message function, and updating function
respectively. GAT's \(AGG\) is linear layer, and \(MES\) is multi-head
attention function. So the attention mechanism in GAT could be described
as follow: \begin{equation}
h_{i}^{\prime}(K)=Pool_{k=1}^{K} h'_{ik}
\end{equation} \begin{equation}
h_{i}^{\prime}=\sigma\left(\sum_{j \in \mathcal{N}_{i}} \alpha_{i j} W h_{j}\right)
\end{equation} \begin{equation}
\alpha_{i,j} =\frac{ \exp\left(\sigma\left(\mathbf{a}^{\top}[\mathbf{\Theta}\mathbf{x}_i \, \Vert \, \mathbf{\Theta}\mathbf{x}_j]\right)\right)}{\sum_{k \in \mathcal{N}(i) \cup \{ i \}}\exp\left(\sigma\left(\mathbf{a}^{\top}[\mathbf{\Theta}\mathbf{x}_i \, \Vert \, \mathbf{\Theta}\mathbf{x}_k]\right)\right)}
\end{equation}

where \(Pool\) denotes pooling function in multi-heads, in the work we
use mean pooling function; \(K\) denotes number of heads;
\(\alpha_{ij}\) denotes normalized attention score of source node \(j\)
on target node \(i\) . \(\sigma\) denotes activation function.

\hypertarget{agdn}{%
\paragraph{3.4.2 AGDN}\label{agdn}}

By denoting transition matrix , parameters matrix, attention vector as
\(\mathbf{T}^{(l)}\), \(\mathbf{W}^{(l)},a_{hw}^{(l)}\) respectively,
AGDN-HA layer could be defined as , \begin{equation}
\mathbf{H}^{(l)}=AGDN(\mathcal{G},\mathbf{H}^{(l-1)},\mathbf{W}^{(l)},a_{hw}^{(l)})
\end{equation} \begin{equation}
\hat{\mathbf{H}}^{(0,l)}=\mathbf{H}^{(l-1)}\mathbf{W}^{(l)}
\end{equation} \begin{equation}
\hat{\mathbf{H}}^{(k,l)}=\mathbf{T}^{(l)}\hat{\mathbf{H}}^{(k-1,l)}
\end{equation}

To be more specific, the output of AGDN-HA layer could be written as:
\begin{equation}
\boldsymbol{H}^{(l)}=\sum_{k=0}^{K} \boldsymbol{\Theta}^{(k, l)} \hat{\boldsymbol{H}}^{(k, l)}+\boldsymbol{H}^{(l-1)} \boldsymbol{W}_{r}^{(l)}
\end{equation} \begin{equation}
\Theta^{(k,l)}=diag(\theta^{(k,l)}_1,...,\theta_N^{(k,l)})
\end{equation} \begin{equation}
\theta_{i}^{(k, l)}=\frac{\exp \left(\sigma\left(\left[\hat{H}_{i}^{(0, l)}|| \hat{H}_{i}^{(k, l)}\right] \cdot a_{h w}^{(l)}\right)\right)}{\sum_{k=0}^{K} \exp \left(\sigma\left(\left[\hat{H}_{i}^{(0, l)} \| \hat{H}_{i}^{(k, l)}\right] \cdot a_{h w}^{(l)}\right)\right)}
\end{equation}

Where \(\hat{H}_i^{(k,l)}\) denotes representation vector of layer
\(l\), aggregation \(k\), node \(i\).

\hypertarget{correct-and-smooth}{%
\subsubsection{3.5 Correct and Smooth}\label{correct-and-smooth}}

The reason why we choose Correct and
Smooth(\protect\hyperlink{ref-huang2020combining}{{[}29{]}}) among large
amouts of label propgation algorithm on GNN is that it has two
advantages: 1), Correct and Smooth is such an effective label
propagation algorithm that even combing with MLP, it outperforms most of
complicate GNN models. 2), Correct and Smooth is a post-processing
module that does not disturb base model, and thanks to that we can
evaluate base model properly. Correct and Smooth contains of two steps:
Correct module corrects the base predictions by modeling correlated
error and Smooth module smooths the soft prediction by correlated error.

\hypertarget{correcting}{%
\paragraph{3.5.1 Correcting}\label{correcting}}

The basic idea of correcting module is that errors in the base
prediction to be positively correlated along edges in the graph, In
other words, an error at node \(i\) increases the chance of a similar
error at neighboring nodes of \(i\). Denoting error matrix as \(E\), we
can get: \begin{equation}
E_{L_t} = Z_{L_t}-Y_{L_t}, E_{L_v}=0, E_{U}=0
\end{equation}

where \(L_t, L_v, U,Z,Y\) denote training labeled data, validating
labeled data, unlabeled data, base predictions, ground-truth labels
repectively.

After that, label spreading technique is used to smooth the err:
\begin{equation}
\hat{E}=\underset{W \in \mathbb{R}^{n \times c}}{\arg \min } \operatorname{trace}\left(W^{T}(I-S) W\right)+\mu\|W-E\|_{F}^{2}
\end{equation}

Where\(S=D^{-1/2}AD^{-1/2}\), \(\mathbf{I}\) denote unit matrix.

Add the solution of \((19)\) to \(Z\) with some scale method to get
corrected prediction: \begin{equation}
Z^{(r)}=f(Z,\hat{E})
\end{equation}

Where \(f\) denotes scale and add operation, here we adapt scaled fixed
diffusion(FDiff-scale), then we get: \begin{equation}
Z^{(r)}=Z+s\hat{E}
\end{equation}

Where \(s\) is a hyperparameter.

\hypertarget{smoothing}{%
\paragraph{3.5.2 Smoothing}\label{smoothing}}

Based on the assumption that adjacent nodes in the graph are likely to
have similar labels, we can smooth \(Z^{(r)}\) to get final predictions.
Start with best guess \(G\): \begin{equation}
G_{L_t}=Y_{L_t},\ \ \ G_{L_v,U}=Z^{(r)}_{L_v,U}
\end{equation}

And implement following iterating until convergence: \begin{equation}
G^{(t+1)}=(1-\alpha)G+\alpha SG^{(t)}
\end{equation}

Where \(G^{(0)}=G\).

The smoothed result then could be used to make final predictions.

\hypertarget{experiments}{%
\subsection{4. Experiments}\label{experiments}}

We conduct detailed experiments on ogbn-arxiv dataset to examine the
effectiveness of distance and hop-wise structure encoding(DHSE) module,
main results are presented below.

\hypertarget{results}{%
\subsubsection{4.1 Results}\label{results}}

\hypertarget{performance-comparison}{%
\paragraph{4.1.1 Performance Comparison}\label{performance-comparison}}

Comparing with baseline models, GATs with DHSE and combining with
Correct and Smooth get significant improvement. It demonstrates that
DHSE provides GATs the high-level multi-hop distance and structure
information that GATs cannot catch with original neighbor-averaging
mechanism.

Table 1 performance comparison

\begin{longtable}[]{@{}lll@{}}
\toprule
model & Valid Accuracy & Test Accuracy \\
\midrule
\endhead
\textbf{AGDN+BoT+self-KD+Correct and Smooth} & 0.7518 ± 0.0009 & 0.7431
± 0.0014 \\
\textbf{UniMP\_v2} & 0.7506 ± 0.0009 & 0.7397 ± 0.0015 \\
\textbf{GAT + Correct and Smooth} & 0.7484 ± 0.0007 & 0.7386 ± 0.0014 \\
\textbf{AGDN (GAT-HA+3\_heads)} & 0.7483 ± 0.0009 & 0.7375 ± 0.0021 \\
\textbf{MLP + Correct and Smooth} & 0.7391 ± 0.0015 & 0.7312 ± 0.0012 \\
\textbf{DHSEGAT+Correct and Smooth} & 0.7441 ± 0.0012 & 0.7425 ±
0.0016 \\
\textbf{DHSEAGDN+Correct and Smooth} & 0.7462 ± 0.0013 & \textbf{0.7439
± 0.0019} \\
\bottomrule
\end{longtable}

\hypertarget{best-performance-of-dhseagdn-and-dhsegat-with-and-without-correct-and-smooth}{%
\paragraph{4.1.2 Best Performance of DHSEAGDN and DHSEGAT with and
without Correct and
Smooth}\label{best-performance-of-dhseagdn-and-dhsegat-with-and-without-correct-and-smooth}}

We argue that DHSE provides GATs the high-level multi-hop distance and
structure information that neighbor-averaging mechanism cannot catch,
however, initial GAT's classifier may neglect the information that
intrinsic features provide with those information. Therefore, the
performance of base model may not significantly improved. But combining
with Correct and Smooth, those information could be effectively used and
the overall performance be improved.

Since DHSEAGDN and DHSEGAT get their best performance in different
hyperparameter setting, so we report the results separately as below.
With same hyperparameter setting, the performance of enhanced models
(ie, DHSEAGDN and AGDN) and base model (ie, AGDN and GAT) are very
close, but with Correct and Smooth, their difference are very
significant.

Table 2 best performance of DHSEAGDN with and without Correct and Smooth

\begin{longtable}[]{@{}lll@{}}
\toprule
model & Valid Accuracy & Test Accuracy \\
\midrule
\endhead
\textbf{DHSEAGDN} & 0.7427 ± 0.0007 & 0.7283 ± 0.0030 \\
\textbf{DHSEAGDN+Correct and Smooth} & 0.7462 ± 0.0013 &
\textbf{\emph{0.7439 ± 0.0019}} \\
\textbf{AGDN} & 0.7429 ± 0.0010 & 0.7297 ± 0.0021 \\
\textbf{AGDN+Correct and Smooth} & 0.7471 ± 0.0007 & 0.7387 ± 0.0018 \\
\bottomrule
\end{longtable}

Table 3 best performance of DHSEGAT with and without Correct and Smooth

\begin{longtable}[]{@{}lll@{}}
\toprule
model & Valid Accuracy & Test Accuracy \\
\midrule
\endhead
\textbf{DHSEGAT} & 0.7412 ± 0.0008 & 0.7261 ± 0.0021 \\
\textbf{DHSEGAT+Correct and Smooth} & 0.7441 ± 0.0012 & \textbf{0.7425 ±
0.0016} \\
\textbf{GAT} & 0.7419 ± 0.0005 & 0.7273 ± 0.0012 \\
\textbf{GAT+Correct and Smooth} & 0.7440 ± 0.0012 & 0.7371 ± 0.0023 \\
\bottomrule
\end{longtable}

\hypertarget{marginal-improvement-with-same-hyperparameter-as-agdn}{%
\paragraph{4.1.3 Marginal Improvement with Same Hyperparameter as
AGDN}\label{marginal-improvement-with-same-hyperparameter-as-agdn}}

To demonstrate the marginal improvement of models with DHSE, we test the
models with same hyperparameter setting as
AGDN(\protect\hyperlink{ref-sun2020adaptive}{{[}23{]}}), the results are
presented as Table 4. We can see that combining with Correct and Smooth,
DHSE can improve the performance of base models significantly.

Table 4 marginal improvement of DHSE

\begin{longtable}[]{@{}lll@{}}
\toprule
model & Valid Accuracy & Test Accuracy \\
\midrule
\endhead
\textbf{DHSEAGDN} & 0.7464 ± 0.0008 & 0.7346 ± 0.0007 \\
\textbf{DHSEAGDN+Correct and Smooth} & 0.7485 ± 0.0008 &
\textbf{\emph{0.7414 ± 0.0010}} \\
\textbf{AGDN} & 0.7432 ± 0.0007 & 0.7344 ± 0.0018 \\
\textbf{AGDN+Correct and Smooth} & 0.7447 ± 0.0009 & 0.7398 ± 0.0010 \\
\textbf{DHSEGAT} & 0.7448 ± 0.0007 & 0.7308 ± 0.0014 \\
\textbf{DHSEGAT+Correct and Smooth} & 0.7471 ± 0.0006 &
\textbf{\emph{0.7388 ± 0.0017}} \\
\textbf{GAT} & 0.7414 ± 0.0005 & 0.7298 ± 0.0022 \\
\textbf{GAT+Correct and Smooth} & 0.7438 ± 0.0006 & 0.7369 ± 0.0018 \\
\bottomrule
\end{longtable}

\hypertarget{ablation-study}{%
\subsubsection{4.2 Ablation Study}\label{ablation-study}}

In this section, we examine the effectiveness of every component of DHSE
in improving performance of base model. DHSE contains of three parts:
encoding layers, structure information and distance distributional
information. The ablation study of the three parts are present as below.

\hypertarget{encoding-layers}{%
\paragraph{4.2.1 Encoding Layers}\label{encoding-layers}}

From Table 5 we can see that, encoding layers are indivisible part of
DHSEGATs. Without encoding layers, their performance are inferior to
base model.

Table 5 DHSEGATs without encoding layers

\begin{longtable}[]{@{}lll@{}}
\toprule
model & Valid Accuracy & Test Accuracy \\
\midrule
\endhead
\textbf{AGDN+raw features} & 0.7398 ± 0.0020 & 0.7303 ± 0.0022 \\
\textbf{AGDN+raw features+Correct and Smooth} & 0.7384 ± 0.0023 & 0.7377
± 0.0026 \\
\textbf{GAT+raw features} & 0.7378 ± 0.0006 & 0.7244 ± 0.0018 \\
\textbf{GAT+raw features+Correct and Smooth} & 0.7407 ± 0.0014 & 0.7374
± 0.0020 \\
\bottomrule
\end{longtable}

\hypertarget{structure-information}{%
\paragraph{4.2.2 Structure Information}\label{structure-information}}

The strucutre information of DHSE contains of three different level:
degree, node-level and graph-level information. From Table 6, Table 7
and Table 8, we can see that degree, node-level structure and
graph-level structure information contribute similar marginal
improvement to DHSEGATs accordingly.

Table 6 DHSEGATs without degree information

\begin{longtable}[]{@{}lll@{}}
\toprule
model & Valid Accuracy & Test Accuracy \\
\midrule
\endhead
\textbf{DHSEAGDN-degree} & 0.7437 ± 0.0005 & 0.7315 ± 0.0013 \\
\textbf{DHSEAGDN-degree+Correct and Smooth} & 0.7443 ± 0.0010 & 0.7417 ±
0.0008 \\
\textbf{DHSEGAT-degree} & 0.7423 ± 0.0005 & 0.7286 ± 0.0021 \\
\textbf{DHSEGAT-degree+Correct and Smooth} & 0.7447 ± 0.0010 & 0.7379 ±
0.0022 \\
\bottomrule
\end{longtable}

Table 7 DHSEGATs without node-level structure information

\begin{longtable}[]{@{}lll@{}}
\toprule
model & Valid Accuracy & Test Accuracy \\
\midrule
\endhead
\textbf{DHSEAGDN-node} & 0.7447 ± 0.0008 & 0.7269 ± 0.0012 \\
\textbf{DHSEAGDN-node+Correct and Smooth} & 0.7474 ± 0.0010 & 0.7425 ±
0.0015 \\
\textbf{DHSEGAT-node} & 0.7438 ± 0.0004 & 0.7301 ± 0.0022 \\
\textbf{DHSEGAT-node+Correct and Smooth} & 0.7450 ± 0.0011 & 0.7402 ±
0.0012 \\
\bottomrule
\end{longtable}

Table 8 DHSEGATs without graph-level structure information

\begin{longtable}[]{@{}lll@{}}
\toprule
model & Valid Accuracy & Test Accuracy \\
\midrule
\endhead
\textbf{DHSEAGDN-graph} & 0.7444 ± 0.0008 & 0.7290 ± 0.0036 \\
\textbf{DHSEAGDN-graph+Correct and Smooth} & 0.7463 ± 0.0016 & 0.7419 ±
0.0028 \\
\textbf{DHSEGAT-graph} & 0.7422 ± 0.0006 & 0.7265 ± 0.0012 \\
\textbf{DHSEGAT-graph+Correct and Smooth} & 0.7450 ± 0.0016 & 0.7399 ±
0.0019 \\
\bottomrule
\end{longtable}

\hypertarget{distance-distributional-information}{%
\paragraph{4.2.3 Distance Distributional
Information}\label{distance-distributional-information}}

From Table 9 we can see that distance distributional information
provides indivisible information to DHSEAGDN, but not to DHSEGAT. We can
see that without distance distributional information, the performance of
DHSEGAT is even improved slightly. We can presume that different model
need different information. For GAT, distance information is not
indivisible, however for AGDN, distance information is indivisible
because of its graph diffusion process.

Table 9 DHSEGATs without distance distributional information

\begin{longtable}[]{@{}lll@{}}
\toprule
model & Valid Accuracy & Test Accuracy \\
\midrule
\endhead
\textbf{DHSEAGDN-distance} & 0.7465 ± 0.0004 & 0.7315 ± 0.0032 \\
\textbf{DHSEAGDN-distance+Correct and Smooth} & 0.7473 ± 0.0014 & 0.7408
± 0.0020 \\
\textbf{DHSEGAT-distance} & 0.7429 ± 0.0006 & 0.7303 ± 0.0017 \\
\textbf{DHSEGAT-distance+Correct and Smooth} & 0.7443 ± 0.0007 & 0.7394
± 0.0020 \\
\bottomrule
\end{longtable}

\hypertarget{conclusion}{%
\subsection{5. Conclusion}\label{conclusion}}

The disadvantage of GNN in catching structure information of graph has
been broadly proven, and the ability of structure and distance
information to improve performance of GNN has been widely proven as
well. Inspired by
Graphormer\protect\hyperlink{ref-ying2021transformers}{{[}7{]}}, we
investigate popular structure indicators` computation complexity and
propose a structure indicator combination with \(O(|N|)\) computation
complexity, and propose a distance distributional information encoding
scheme. We propose a distance and hop-wise structure information
injecting scheme which fits universal GNN models. Detailed experiments
shows its ability on improving the expressive power of GNNs.

By detailed experiments, we demonstrate that label information that
Correct and Smooth provides is indivisible for DHSE on improving the
expressive power of GNNs. Without Correct and Smooth, DHSE can only
slightly improve the performance of GATs. However, with Correct and
Smooth, DHSE can significantly improve the performance. It implies that
DHSE provides GATs the information that neighbor-averaging schema cannot
provide, and neighbor-averaging schema cannot effectively use those
information but Correct and Smooth can. Moreover, ablation study shows
that encoding layers are key compoments of DHSEGATs in improving
performance of base model, and every structure information make similar
contribution, while distance distributional information is indisvisible
part of DHSE.

\hypertarget{reference}{%
\subsection*{Reference}\label{reference}}
\addcontentsline{toc}{subsection}{Reference}

\hypertarget{refs}{}
\begin{CSLReferences}{0}{0}
\leavevmode\vadjust pre{\hypertarget{ref-li2018deeper}{}}%
\CSLLeftMargin{{[}1{]} }
\CSLRightInline{Q. Li, Z. Han, and X.-M. Wu, {``Deeper insights into
graph convolutional networks for semi-supervised learning,''} in
\emph{Thirty-{Second AAAI} conference on artificial intelligence}, 2018.
}

\leavevmode\vadjust pre{\hypertarget{ref-oono2019graph}{}}%
\CSLLeftMargin{{[}2{]} }
\CSLRightInline{K. Oono and T. Suzuki, {``Graph neural networks
exponentially lose expressive power for node classification,''} 2019.
{[}Online{]}. Available: \url{https://arxiv.org/abs/1905.10947}}

\leavevmode\vadjust pre{\hypertarget{ref-dehmamy2019understanding}{}}%
\CSLLeftMargin{{[}3{]} }
\CSLRightInline{N. Dehmamy, A.-L. Barabási, and R. Yu, {``Understanding
the representation power of graph neural networks in learning graph
topology,''} 2019. {[}Online{]}. Available:
\url{https://arxiv.org/abs/1907.05008}}

\leavevmode\vadjust pre{\hypertarget{ref-monti2018motifnet}{}}%
\CSLLeftMargin{{[}4{]} }
\CSLRightInline{F. Monti, K. Otness, and M. M. Bronstein, {``Motifnet: A
motif-based graph convolutional network for directed graphs,''} in
\emph{2018 {IEEE Data Science Workshop} ({DSW})}, 2018, pp. 225--228. }

\leavevmode\vadjust pre{\hypertarget{ref-bouritsas2020improving}{}}%
\CSLLeftMargin{{[}5{]} }
\CSLRightInline{G. Bouritsas, F. Frasca, S. Zafeiriou, and M. M.
Bronstein, {``Improving graph neural network expressivity via subgraph
isomorphism counting,''} 2020. {[}Online{]}. Available:
\url{https://arxiv.org/abs/2006.09252}}

\leavevmode\vadjust pre{\hypertarget{ref-you2021identityaware}{}}%
\CSLLeftMargin{{[}6{]} }
\CSLRightInline{J. You, J. Gomes-Selman, R. Ying, and J. Leskovec,
{``Identity-aware graph neural networks,''} 2021. {[}Online{]}.
Available: \url{https://arxiv.org/abs/2101.10320}}

\leavevmode\vadjust pre{\hypertarget{ref-ying2021transformers}{}}%
\CSLLeftMargin{{[}7{]} }
\CSLRightInline{C. Ying \emph{et al.}, {``Do {Transformers Really
Perform Bad} for {Graph Representation}?''} 2021. {[}Online{]}.
Available: \url{https://arxiv.org/abs/2106.05234}}

\leavevmode\vadjust pre{\hypertarget{ref-you2019positionaware}{}}%
\CSLLeftMargin{{[}8{]} }
\CSLRightInline{J. You, R. Ying, and J. Leskovec, {``Position-aware
graph neural networks,''} in \emph{International {Conference} on
{Machine Learning}}, 2019, pp. 7134--7143. }

\leavevmode\vadjust pre{\hypertarget{ref-li2020distance}{}}%
\CSLLeftMargin{{[}9{]} }
\CSLRightInline{P. Li, Y. Wang, H. Wang, and J. Leskovec, {``Distance
encoding: Design provably more powerful neural networks for graph
representation learning,''} 2020. {[}Online{]}. Available:
\url{https://arxiv.org/abs/2009.00142}}

\leavevmode\vadjust pre{\hypertarget{ref-alsentzer2020subgraph}{}}%
\CSLLeftMargin{{[}10{]} }
\CSLRightInline{E. Alsentzer, S. G. Finlayson, M. M. Li, and M. Zitnik,
{``Subgraph neural networks,''} 2020. {[}Online{]}. Available:
\url{https://arxiv.org/abs/2006.10538}}

\leavevmode\vadjust pre{\hypertarget{ref-dasoulas2019coloring}{}}%
\CSLLeftMargin{{[}11{]} }
\CSLRightInline{G. Dasoulas, L. D. Santos, K. Scaman, and A. Virmaux,
{``Coloring graph neural networks for node disambiguation,''} 2019.
{[}Online{]}. Available: \url{https://arxiv.org/abs/1912.06058}}

\leavevmode\vadjust pre{\hypertarget{ref-xu2018how}{}}%
\CSLLeftMargin{{[}12{]} }
\CSLRightInline{K. Xu, W. Hu, J. Leskovec, and S. Jegelka, {``How
powerful are graph neural networks?''} 2018. {[}Online{]}. Available:
\url{https://arxiv.org/abs/1810.00826}}

\leavevmode\vadjust pre{\hypertarget{ref-arvind2020weisfeilerleman}{}}%
\CSLLeftMargin{{[}13{]} }
\CSLRightInline{V. Arvind, F. Fuhlbrück, J. Köbler, and O. Verbitsky,
{``On weisfeiler-leman invariance: Subgraph counts and related graph
properties,''} \emph{Journal of Computer and System Sciences}, vol. 113,
pp. 42--59, 2020. }

\leavevmode\vadjust pre{\hypertarget{ref-chen2020can}{}}%
\CSLLeftMargin{{[}14{]} }
\CSLRightInline{Z. Chen, L. Chen, S. Villar, and J. Bruna, {``Can graph
neural networks count substructures?''} 2020. {[}Online{]}. Available:
\url{https://arxiv.org/abs/2002.04025}}

\leavevmode\vadjust pre{\hypertarget{ref-vignac2020building}{}}%
\CSLLeftMargin{{[}15{]} }
\CSLRightInline{C. Vignac, A. Loukas, and P. Frossard, {``Building
powerful and equivariant graph neural networks with structural
message-passing,''} 2020. {[}Online{]}. Available:
\url{https://arxiv.org/abs/2006.15107}}

\leavevmode\vadjust pre{\hypertarget{ref-nt2019revisiting}{}}%
\CSLLeftMargin{{[}16{]} }
\CSLRightInline{H. Nt and T. Maehara, {``Revisiting graph neural
networks: All we have is low-pass filters,''} 2019. {[}Online{]}.
Available: \url{https://arxiv.org/abs/1905.09550}}

\leavevmode\vadjust pre{\hypertarget{ref-frasca2020sign}{}}%
\CSLLeftMargin{{[}17{]} }
\CSLRightInline{F. Frasca, E. Rossi, D. Eynard, B. Chamberlain, M.
Bronstein, and F. Monti, {``Sign: Scalable inception graph neural
networks,''} 2020. {[}Online{]}. Available:
\url{https://arxiv.org/abs/2004.11198}}

\leavevmode\vadjust pre{\hypertarget{ref-sun2021scalable}{}}%
\CSLLeftMargin{{[}18{]} }
\CSLRightInline{C. Sun and G. Wu, {``Scalable and {Adaptive Graph Neural
Networks} with {Self}-{Label}-{Enhanced} training,''} 2021.
{[}Online{]}. Available: \url{https://arxiv.org/abs/2104.09376}}

\leavevmode\vadjust pre{\hypertarget{ref-velickovic2017graph}{}}%
\CSLLeftMargin{{[}19{]} }
\CSLRightInline{P. Veličković, G. Cucurull, A. Casanova, A. Romero, P.
Lio, and Y. Bengio, {``Graph attention networks,''} 2017. {[}Online{]}.
Available: \url{https://arxiv.org/abs/1710.10903}}

\leavevmode\vadjust pre{\hypertarget{ref-wu2020graph}{}}%
\CSLLeftMargin{{[}20{]} }
\CSLRightInline{T. Wu, H. Ren, P. Li, and J. Leskovec, {``Graph
information bottleneck,''} 2020. {[}Online{]}. Available:
\url{https://arxiv.org/abs/2010.12811}}

\leavevmode\vadjust pre{\hypertarget{ref-hu2020gptgnn}{}}%
\CSLLeftMargin{{[}21{]} }
\CSLRightInline{Z. Hu, Y. Dong, K. Wang, K.-W. Chang, and Y. Sun,
{``Gpt-gnn: Generative pre-training of graph neural networks,''} in
\emph{Proceedings of the 26th {ACM SIGKDD International Conference} on
{Knowledge Discovery} \& {Data Mining}}, 2020, pp. 1857--1867. }

\leavevmode\vadjust pre{\hypertarget{ref-hu2020heterogeneous}{}}%
\CSLLeftMargin{{[}22{]} }
\CSLRightInline{Z. Hu, Y. Dong, K. Wang, and Y. Sun, {``Heterogeneous
graph transformer,''} in \emph{Proceedings of {The Web Conference}
2020}, 2020, pp. 2704--2710. }

\leavevmode\vadjust pre{\hypertarget{ref-sun2020adaptive}{}}%
\CSLLeftMargin{{[}23{]} }
\CSLRightInline{C. Sun and G. Wu, {``Adaptive graph diffusion networks
with hop-wise attention,''} 2020. {[}Online{]}. Available:
\url{https://arxiv.org/abs/2012.15024}}

\leavevmode\vadjust pre{\hypertarget{ref-wang2020unifying}{}}%
\CSLLeftMargin{{[}24{]} }
\CSLRightInline{H. Wang and J. Leskovec, {``Unifying graph convolutional
neural networks and label propagation,''} 2020. {[}Online{]}. Available:
\url{https://arxiv.org/abs/2002.06755}}

\leavevmode\vadjust pre{\hypertarget{ref-qu2019gmnn}{}}%
\CSLLeftMargin{{[}25{]} }
\CSLRightInline{M. Qu, Y. Bengio, and J. Tang, {``Gmnn: Graph markov
neural networks,''} in \emph{International conference on machine
learning}, 2019, pp. 5241--5250. }

\leavevmode\vadjust pre{\hypertarget{ref-gao2019conditional}{}}%
\CSLLeftMargin{{[}26{]} }
\CSLRightInline{H. Gao, J. Pei, and H. Huang, {``Conditional random
field enhanced graph convolutional neural networks,''} in
\emph{Proceedings of the 25th {ACM SIGKDD International Conference} on
{Knowledge Discovery} \& {Data Mining}}, 2019, pp. 276--284. }

\leavevmode\vadjust pre{\hypertarget{ref-zheng2021gipa}{}}%
\CSLLeftMargin{{[}27{]} }
\CSLRightInline{Q. Zheng \emph{et al.}, {``{GIPA}: General {Information
Propagation Algorithm} for {Graph Learning},''} 2021. {[}Online{]}.
Available: \url{https://arxiv.org/abs/2105.06035}}

\leavevmode\vadjust pre{\hypertarget{ref-klicpera2018predict}{}}%
\CSLLeftMargin{{[}28{]} }
\CSLRightInline{J. Klicpera, A. Bojchevski, and S. Günnemann, {``Predict
then propagate: Graph neural networks meet personalized pagerank,''}
2018. {[}Online{]}. Available: \url{https://arxiv.org/abs/1810.05997}}

\leavevmode\vadjust pre{\hypertarget{ref-huang2020combining}{}}%
\CSLLeftMargin{{[}29{]} }
\CSLRightInline{Q. Huang, H. He, A. Singh, S.-N. Lim, and A. R. Benson,
{``Combining label propagation and simple models out-performs graph
neural networks,''} 2020. {[}Online{]}. Available:
\url{https://arxiv.org/abs/2010.13993}}

\end{CSLReferences}

\end{document}